%% file: acl.tex
\pdfoutput=1

\documentclass[11pt]{article}

\usepackage{acl}

\usepackage{times}
\usepackage{latexsym}
\usepackage{xurl}
\usepackage{enumitem}
\usepackage{booktabs}
\usepackage{multirow}
\usepackage{array}
\usepackage{graphicx}
\usepackage{amsmath}
\usepackage{amsfonts}
\usepackage{subcaption}
\usepackage{makecell}
\usepackage{color,soul}
\usepackage{algorithm}
\usepackage[noend]{algpseudocode}
\usepackage{xcolor,colortbl}
\usepackage{xspace}
\usepackage{tablefootnote}

\usepackage{etoolbox}
\AtBeginEnvironment{quote}{\par\singlespacing\small}
\definecolor{green}{rgb}{0.1,0.1,0.1}
\definecolor{gitgreen}{HTML}{006400}
\definecolor{chocolate}{HTML}{D2691E}
\definecolor{maroon}{HTML}{A00000}
\definecolor{indigo}{HTML}{4B0082}
\definecolor{green}{HTML}{008000}

\usepackage{amssymb}%
\usepackage{pifont}%

\newcolumntype{L}[1]{>{\PreserveBackslash\raggedright}p{#1}}
\newcolumntype{R}[1]{>{\raggedleft\let\newline\\\arraybackslash\hspace{0pt}}m{#1}}

\newcommand{\velocity}[1]{
}
\newcommand{\hanna}[1]{
}
\newcommand{\luke}[1]{
}
\newcommand{\sewon}[1]{
}
\newcommand{\ignore}[1]{}

\newcommand{\NLITrack}{\textsc{gold-comment} track}

\newcommand{\fp}{\texttt{FP}}
\newcommand{\n}{\texttt{N}}

\usepackage{adjustbox}

\newcommand{\Q}{\colorbox{blue!30}{\textbf{\texttt{Q}}}}
\newcommand{\C}{\colorbox{pink!70}{\textbf{\texttt{C}}}}
\newcommand{\FP}{\colorbox{purple!30}{{\textbf{\texttt{FP}}}}}

\newcommand{\myhighlight}[1]{\colorbox{purple!30}{{\textbf{\texttt{#1}}}}}
\newcommand{\mycomment}{\colorbox{pink!70}{\textbf{\texttt{comment}}}}
\newcommand{\myhighlightblue}[1]{\colorbox{green!30}{\textbf{\texttt{#1}}}}

\newcommand{\dataname}{\textsc{Crepe}}
\newcommand{\ourURL}{
\href{https://github.com/velocityCavalry/CREPE}{\nolinkurl{github.com/velocityCavalry/CREPE}}}

\usepackage[T1]{fontenc}

\usepackage[utf8]{inputenc}

\usepackage{microtype}

\title{
    \dataname: Open-Domain Question Answering with False Presuppositions
}

\author{
    Xinyan Velocity Yu$^\dagger$ \quad Sewon Min$^\dagger$ \quad 
    Luke Zettlemoyer$^\dagger$ \quad Hannaneh Hajishirzi$^{\dagger,\ddagger}$
    \\
    $^\dagger$University of Washington \quad $^\ddagger$Allen Institute for Artificial Intelligence \\
    \texttt{ \{xyu530,sewon,lsz,hannaneh\}@cs.washington.edu} \\ }

\begin{document}
\maketitle
\begin{abstract}
    Information seeking users often pose questions with false presuppositions, especially when asking about unfamiliar topics. Most existing question answering (QA) datasets, in contrast, assume all questions have well defined answers. We introduce \dataname, a QA dataset containing a natural distribution of presupposition failures from online information-seeking forums. We find that 25\% of questions contain false presuppositions, and provide annotations for these presuppositions and their corrections.
    Through extensive baseline experiments, we show that adaptations of existing open-domain QA models can find presuppositions moderately well, but struggle when predicting whether a presupposition is factually correct. This is in large part due to difficulty in retrieving relevant evidence passages from a large text corpus. %
    \dataname\ provides a benchmark to study question answering in the wild, and our analyses provide avenues for future work in better modeling and further studying the task. %
\end{abstract}

\section{Introduction}\label{sec:intro}\input{sections/01-intro}
\section{Background}\label{sec:related}\input{sections/02-related}

\section{Dataset: \dataname}\label{sec:data}\input{sections/03-data}
\section{Task Setup}\label{sec:task}\input{sections/04-task-setup}
\section{Experiments: Detection}\label{sec:exp-detection}\input{sections/05-exp-detection}
\section{Experiments: Writing}\label{sec:exp-writing}\input{sections/06-exp-writing}

\section{Discussion}\label{sec:discuss}\input{sections/07-discuss}
\section{Conclusion}\label{sec:concl}\input{sections/08-concl}

\section*{Acknowledgements}
We thank Daniel Fried and Julian Michael for their feedback on an early stage of the project. We thank Zeqiu Wu and H2Lab members for comments on the paper.
We thank Li Du for suggesting the name of the dataset.

\bibliography{acl,abbr,custom}
\bibliographystyle{acl_natbib}

\clearpage
\appendix
\input{sections/10-app}

\end{document}

%% file: sections/01-intro.tex
\begin{figure}[t]
\resizebox{1.02\columnwidth}{!}{\includegraphics[trim={0cm 7cm 0cm 0cm},clip]{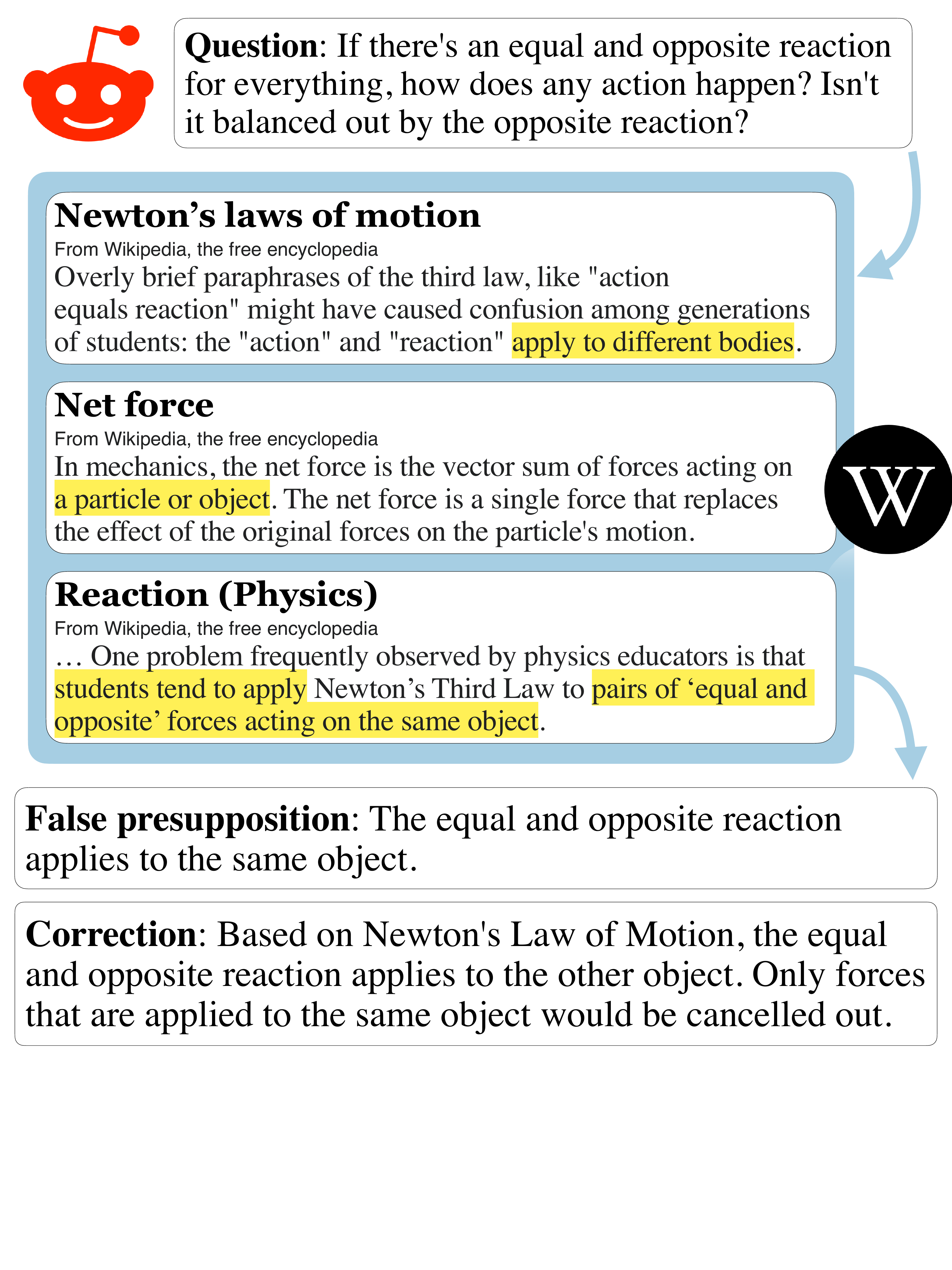}}\vspace{-.1cm}
\caption{
    An example question written by an online user that contains a false presupposition.
    The model is required to (a) identify the presupposition made in the question that is false based on world knowledge, and (b) write the correction.
    We also show three paragraphs from the English Wikipedia that are evidence paragraphs.%
}\label{fig:teaser}
\end{figure}

When an information-seeking user poses a question about the topic they are unfamiliar with, they can often introduce false presuppositions~\citep{kaplan-1978-indirect-responses,duvzi2015questions} which are assumed but not directly stated. 
For instance, the question in Figure~\ref{fig:teaser} incorrectly presupposes that the equal and opposite reactions in Newton's law apply to the same object.
Although such a question is unanswerable, we might still hope to identify the confusion and explain it to the user. 

However, this functionality goes well beyond prior open-domain QA task formulations, which focus on questions with a valid direct answer~\citep{rajpurkar-etal-2016-squad, kwiatkowski-etal-2019-natural} or that are unanswerable from lack of evidence~\citep{rajpurkar-etal-2018-know,choi2018quac,asai-choi-2021-challenges}.
While recent work studies unverifiable presuppositions in reading-comprehension-style questions given evidence context~\citep{kim-etal-2021-linguist}, there has been no work that identifies and corrects a presupposition that is false based on {\em global, factual} knowledge.

In this paper, we introduce \textbf{\dataname} (\textsc{\textbf{C}}or\textsc{\textbf{re}}ction of \textsc{\textbf{p}}r\textsc{\textbf{e}}supposition), a new dataset consisting of 8,400 Reddit questions with (1) whether there is any false presuppositions annotated, and (2) if any, the presupposition and its correction written.
We find 25\% of questions posted on Reddit~\citep{fan-etal-2019-eli5} include false presuppositions, where the best response is to provide a correction of the presuppositions.
A similar portion of questions have false presuppositions in the domain of research papers~\citep{dasigi-etal-2021-dataset}, scientific reviews~\citep{kang18naacl}, and social media~\citep{sap-etal-2020-social}, which we discuss in Section~\ref{sec:discuss}.

While the \dataname\ annotation task is challenging due to the need for extensive background knowledge and inherent debatability, we leverage the {\em most upvoted comments} written by community users to efficiently annotate the data.
Intuitively, the most upvoted comments are likely to be factually correct, and typically also identify and correct any false presuppositions made in the question.
By designing an annotation pipeline using these comments, we were able to collect high-quality data with relatively low cost.
Our data analysis (Table~\ref{tab:data-analysis}) shows that the types of false presuppositions are diverse, ranging from relatively explicit presuppositions (e.g., false clauses or false predicate) to subtle, nuanced presuppositions (e.g., false causality or false existential presuppositions).

We define two tracks with varying levels of difficulty, and introduce models to set baseline performance levels for each.
A model is given either the question only (the main track) or the question and the comment (the \NLITrack), and is supposed to perform two subtasks:
identification of whether or not there is a false presupposition (the \textbf{detection} subtask) and generation of presuppositions and their corrections (the \textbf{writing} subtask).
For the writing subtask,
in additional to the automatic evaluation,
we propose a systematic human evaluation scheme based on~\citet{celikyilmaz2020evaluation} that considers fluency, correctness (precision of the information), adequacy (recall of the information) and consistency.

We include a range of baselines, including a question-only model, a nearest-neighbor model, and a competitive model based on the state-of-the-art passage retrieval~\citep{krishna-etal-2021-hurdles} and pretrained language models~\citep{liu2019roberta,2020t5}.
Results and analyses indicate that
(1) retrieval is very challenging since simply retrieving passages in the topic of the question is not enough;
(2) models do moderately well in identifying explicit false presuppositions,
and
(3) models struggle with identifying implicit presuppositions, and explaining how and why the presupposition is false.
We also discuss open problems, such as an inherent ambiguity in the validity of presuppositions and inconsistency between different websites.
The data, baselines, and the evaluation script are available at \ourURL. %

%% file: sections/02-related.tex
\subsection{Question Answering}\label{subsec:background-qa}
There has been significant work on question answering, where the model receives a natural language, open-domain question and is required to return a short, concise answer~\citep{voorhees-tice-2000-trec,lee-etal-2019-latent}.
Most work focuses on questions that have a short text span as a correct answer. Other work studies unanswerable questions, but they study questions that are either intentionally written to be unanswerable~\citep{rajpurkar-etal-2018-know}, or where there is a lack of evidence to find the answer~\citep{choi2018quac,asai-choi-2021-challenges}.
More recently, \citet{kim-etal-2021-linguist} studies unverifiable presuppositions in questions under the given context, but using questions from \citet{kwiatkowski-etal-2019-natural} whose presuppositions are mostly not false based on {\em global} knowledge.\footnote{
Less than 5\% of questions from \citet{kwiatkowski-etal-2019-natural} contains false presupposition under our definition, likely because their questions are aggressively filtered.
}

In this work, we focus on open-domain questions with presuppositions that are false based on global knowledge. They are categorized to be unanswerable based on the taxonomy from previous work, but we argue they can be {\em answered} by providing false presuppositions and their corrections.
We show that false presuppositions in questions are \textbf{prevalent}. 25\% of questions contain false presuppositions in the domain of online forums (for which we collect annotations), research papers~\citep{dasigi-etal-2021-dataset}, scientific reviews~\citep{kang18naacl}, and social media~\citep{sap-etal-2020-social}.

\subsection{Presupposition}\label{subsec:background-presupposition}
Under a {\em pragmatic} point of view, a presupposition is a condition that a speaker would normally expect to hold in the common ground between discourse participants when that sentence is uttered~\citep{sep-presupposition,stalnaker1977pragmatic}.
Unlike semantic presuppositions~\citep{strawson1950referring}, pragmatic presuppositions cannot easily be traced to specific words or phrases, but rather depend on the context and the expectations of the discourse participants~\citep{potts2015presupposition}.
A key property of pragmatic presuppositions is  that they are {\em backgrounded}---a pragmatic property of being a meaning that the speaker presumes to be mutual public knowledge.
Whether or not one is a presupposition is inherently debatable.
In this work, we carefully define a false presupposition in the question by leveraging the most voted comment in the web community, as we discuss further in Section~\ref{subsec:data-annotation}.

False presuppositions in {\em questions} have been discussed in the linguistic literature~\citep{kaplan-1978-indirect-responses, duvzi2015questions}. A key property of false presuppositions in the question is that they make a question infelicitous because there is no direct answer. \citet{kaplan-1978-indirect-responses} claims that an adequate and unambiguous answer to such a question is a negated presupposition, referred to as corrective indirect response. We follow them in providing the negated presupposition as the response to the question, and build the first benchmark based on questions written by information-seeking users.

%% file: sections/03-data.tex
\begin{table*}[!t]
    \centering \footnotesize
    \begin{tabular}{lrrrrrrr}
        \toprule
            \multirow{2}{*}{Data split} & 
            \multicolumn{2}{c}{\# Questions} & \multicolumn{4}{c}{\# Tokens} & \multirow{2}{*}{Posting time} \\
            \cmidrule(lr){2-3} \cmidrule(lr){4-7}
            & Tot & w/ FP & Q & PS & CR & CM \\
        \midrule
            Training & 3,462 & 907 (26.2\%)        & 15.6 & 10.3 & 16.5 & 95.2  & 2018 \\
            Development &   2,000 & 544 (27.2\%) & 16.1 & 10.3 & 15.6 & 91.0 & Jan--Jun 2019\\
            Test &          3,004 & 751 (25.0\%) & 16.4 & 11.8 & 16.8 & 92.5  & Jul--Dec 2019 \\
        \midrule
            Unlabeled training & 196,385 & - & 15.7 & - & - & 96.6  & 2011--2018 \\
        \midrule
            Total (labeled only) & 8,466 & 2,202 (26.0\%) & 16.0 & 10.8 & 16.5 & 93.3 &    \\
            Total (labeled+unlabeled) &  204,851 &  - & 15.7 & - & - & 96.5 &   \\
        \bottomrule
    \end{tabular}\vspace{-.1em}
    \caption{Data statistics. 
    {\em \# Questions} indicate the number of questions in total and with false presupposition.
    {\em \# Token} indicate the number of tokens (based on the whitespace tokenization) in the question (Q), the presupposition (PS), the correction (CR), and the comment (CM). Note that PS and CR are those to be written by the model; CM is given to the model in the \NLITrack.
    Details of tracks are provided in Section~\ref{sec:task}. }\label{tab:data-statistics}
\end{table*}

Our task requires the model to be given a question $q$ and to (a) identify whether $q$ has any false presuppositions, and (b) if yes, generate the false presupposition as well as the correction.

We have three criteria in constructing the dataset:\vspace{-.3em}
\begin{enumerate}\itemsep -.1em
    \item[\textbf{C1.}] Naturalness of the questions: whether the questions in the data are natural questions written by real, information-seeking users.
    \item[\textbf{C2.}] Validity of the presupposition: whether the identified presupposition is highly likely made by the question writer.
    \item[\textbf{C3.}] Correctness and adequacy of the information: whether the information provided in the correction is factually correct (high precision) and adequate to convince the question writer (high recall).
\end{enumerate}

We first describe the source of questions (Section~\ref{subsec:data-source}) which address \textbf{C1}.
We then describe the process of annotation (Section~\ref{subsec:data-annotation}) that addresses \textbf{C2} and \textbf{C3}.
Finally, we present detailed, qualitative analysis of the data (Section~\ref{subsec:data-analysis}).
The formal definition of the task and metrics are provided in Section~\ref{sec:task}.

\subsection{Data Source}\label{subsec:data-source}
Our highest priority is to study false presuppositions that {\em naturally} occur from information-seeking users. While it is significantly easier to manually write questions that would have false presuppositions, 
we think these questions will significantly be different from naturally occurring questions.

Following \citet{fan-etal-2019-eli5}, we use questions posted on the ELI5 subreddit.\footnote{\href{https://www.reddit.com/r/explainlikeimfive}{\nolinkurl{www.reddit.com/r/explainlikeimfive}}}
We made a few modifications to the procedure that \citet{fan-etal-2019-eli5} took in order to improve the data quality.
We first filter questions and comments based on upvotes with a higher threshold. We then split the training, the development and the test data based on the time of the posting: questions on the training set are posted in 2011--2018, questions on the development set are posted in Jan--Jun of 2019, and questions on the test set are posted in Jul--Dec of 2019. We filter out the small subset of questions that include toxic language. Appendix~\ref{app:data-source-details} provides more details. %

\citet{krishna-etal-2021-hurdles} raised a concern that a significant amount of test are duplicates of those on the training set. We provide a detailed analysis in Appendix~\ref{app:data-source-details}. In summary, we think (1) the amount of duplicated (or paraphrased) questions is significantly less than their estimate, %
especially with respect to underlying presuppositions in the questions, and (2) even if there are paraphrased questions, data split based on the time frame is justified based on the real-world scenario.\footnote{We think having similar questions is an inherent property of questions on the web, and the model should be allowed to take whichever approach that is plausible, including the nearest neighbor approach~\citep{lewis-etal-2021-question}.}

\subsection{Data Annotation}\label{subsec:data-annotation}

Meeting the criteria \textbf{C2} and \textbf{C3} can be very difficult for the following reasons:\vspace{-.3em}
\begin{itemize}[leftmargin=1em]\itemsep -.1em
    \item For \textbf{C2}: The validity of presupposition is inherently debatable and largely depends on the background of individuals (Section~\ref{subsec:background-presupposition}).\footnote{
        This is also the case in previous work---for instance, the data annotated by experts in formal semantics and pragmatics can have low agreement~\citep{jeretic-etal-2020-natural}.
    }
    \item For \textbf{C3}: The open-domain nature of the task requires the search of world knowledge on the web, which is extremely expensive and may not be exhaustive enough despite the best efforts made by annotators, as discussed in \citet{kwiatkowski-etal-2019-natural,min-etal-2020-ambigqa}.
\end{itemize}

In this work, we make use of the \textbf{most upvoted comments} written by community users. 
The comment, often written by domain experts, provides a response to the question in the ELI5 subreddit, and has been used as a credible source in prior work~\citep{fan-etal-2019-eli5}.
If the comment identifying a false presupposition has the most upvotes, it is likely that the presupposition is valid (made by a question writer) based on the background context shared by community users, thus satisfying \textbf{C2}. %
Moreover, the comment (1) is highly likely to contain information that is correct and adequate (satisfying \textbf{C3}), and (2) removes the need for exhaustively searching over the web (reducing the annotation cost).

\paragraph{Annotation task.} Annotators are given a pair of the question and the most voted comment, and perform the following steps.\vspace{-.35em}
\begin{enumerate}\itemsep -.15em
    \item Filter out questions that are subjective, are uninformative, or rely too much on personal experience.
    \item Judge whether there is a false presupposition in the question, %
    identified by the comment.
    \item If there is a false presupposition, write the presupposition and a correction as a concise, declarative sentence.
\end{enumerate}
\paragraph{Annotation pipeline.}
We maintain a pool of qualified annotators who passed our qualification task.
We assign two annotators per question, where each annotators independently annotate the question.
We filter out questions if either of the  annotators mark them as such.
If the annotators agreed to on the label (whether or not there is a false presupposition), their label as well as their writings are taken as gold references.
When they disagreed, we assign a third annotators and take a majority vote over three workers.
We find that the disagreements %
are mainly due to inherent ambiguities of the task due to the different interpretation of the question or the comment, or difference in individual background knowledge. We discuss these cases in Section~\ref{subsec:data-analysis}.

\vspace{.1em}
More details in instructions and quality control are provided in Appendix~\ref{app:data-annotation-details}.

\begin{table*}[!t]
    \centering \footnotesize
    \setlength{\tabcolsep}{0.4em}
    \begin{tabular}{p{0.46\textwidth}}
        \toprule
            \multicolumn{1}{c}{\textbf{False clauses (14\%)}} \vspace{.1em} \\
            \textbf{\Q} If water has to be 100 to become steam, how come you don't get heavily burned in saunas?\\
            \textbf{\C} What we often call steam is just water vapor that has started to become visible due to different temperatures of the water vs air. It can exist at many temperatures. (...) \\
            \textbf{\FP} Water has to be 100 degrees to be steam. \\
            \textbf{\Q} Why is the air cold when you fan yourself on a very hot day? Doesn't motion generate heat? \\
            \textbf{\C} Moving air takes away heat faster from your skin. That's why it feels cool. (...) I assume you mean friction generates heat. But the air speed generated by a fan is way too slow to generate any friction heat. \\
            \textbf{\FP} Motion of the fan generates heat. \\
        \midrule
            \multicolumn{1}{c}{\textbf{False properties (22\%)}} \vspace{.1em} \\
            \Q\ How do martial artists who karate chop or punch a cement block not break their hand? \\
            \C\ It's a trick, the blocks are not very strong, and they are being punched or kicked in their weakest points. \\
            \FP\ Chops or cement blocks are strong. \\
            \textbf{\Q} How does your phone tell the difference between a step and random movement? \\
            \textbf{\C} You might be disappointed by this answer, but most of the time, you're not moving your phone so abruptly that the change in acceleration to it is anywhere near the shock that comes %
            when you walk. \\
            \textbf{\FP} A random movement is detectable by a phone. \\
        \midrule
            \multicolumn{1}{c}{\textbf{False existential presupposition (6\%)}} \vspace{.1em} \\
            \textbf{\Q} What uses the space on a hard disk that we're unable to use? For example in a 1TB hard disk, we get about 930GB of usable memory, what happens to the other 70GB? \\
            \textbf{\C} There are %
            TB (terabyte) and TiB (tebibyte). the "ra" ones are using multiplies of 1000. the "bi" ones are using multiplies of 1024. I will do some math for you: 1 TB=1000$^4$B = (...) = 0.93 TiB. There goes your 70 GiB.\\
            \textbf{\FP} In a 1TB hard disk, 70GB is unusable. \\
        \bottomrule
    \end{tabular}\vspace{-.1em}
    \begin{tabular}{p{0.47\textwidth}}
        \toprule
            \multicolumn{1}{c}{\textbf{False predicate (30\%)}} \vspace{.12em} \\
            \textbf{\Q} How exactly is current stored in power plants? 
            \\ 
            \textbf{\C} It's not being stored at all. %
            The power grid is a carefully balanced dance of supply and demand. (...) \\
            \textbf{\FP} Current is stored in power plants. \\
            \textbf{\Q} How can light bounce off objects but it has no mass. \\
            \textbf{\C} Light does not ``bounce''. A photon of light is absorbed by an atom, then (...) \\
            \textbf{\FP} Light bounces off objects. \\
        \midrule
            \multicolumn{1}{c}{\textbf{False (causal) relationship between facts (22\%)}} \vspace{.1em}
            \\
            \textbf{\Q} If there's an equal and opposite reaction for everything, how does any action happen? Isn't it balanced out by the opposite reaction? \\
            \textbf{\C} I don't think you are fully comprehending what `equal' means in this situation. (...) These forces are acting on different bodies so they do not cancel each other out. (...) \\
            \textbf{\FP} The equal and opposite reaction applies to the same object.\\
            \textbf{\Q} In today’s high tech world, how come we are not able to reproduce ancient crafting methods like Roman Concrete, Damascus Steel, or Greek Fire? \\
            \textbf{\C} It's not that we can't reproduce them technologically, it's that the exact method or recipe was lost to history (...) \\
            \textbf{\FP} Ancient crafting methods are not reproducible due to lack of technologies. \\
        \midrule
            \multicolumn{1}{c}{\textbf{Exceptions (4\%) }} \vspace{.1em} \\
            \textbf{\Q} How do bugs and other insects survive winter when they have such a short lifespan? \\
            \textbf{\C} Depends on the insect, some don't have that short of a lifespan. But mostly (...) \\
            \textbf{\FP} (All) insects have a short lifespan. \\
        \midrule
            \multicolumn{1}{c}{\textbf{No false presupposition / Annotation error (2\%)}} \vspace{.1em} \\
            \textbf{\Q} Why can't we artificially make plastic decompose faster? \\
            \textbf{\C} Decomposition happens when something eats the thing that is decomposing. (...) %
            \\
        \bottomrule
    \end{tabular}\vspace{-.1em}
    \caption{Breakdown of types of false presuppositions, based on 50 random samples on the development data. \Q, \C\ and \FP\ indicate the question, the comment, and the presupposition, respectively.
    }\label{tab:data-analysis}
\end{table*}

\subsection{Data Analysis}\label{subsec:data-analysis}

The data statistics are provided in Table~\ref{tab:data-statistics}. We find that over 25\% of questions posted on Reddit includes false presuppositions.\footnote{
    \citet{xu-etal-2022-answer}, who also used the ELI5 subreddit, identified 10\% of questions have false presuppositions (rejected presuppositions) and excluded them in their data.
    The reason their estimate is significantly lower than ours is that they did not include partial rejection while we do.
}

\paragraph{Categorization of false presuppositions.}
We randomly sample 50 questions with false presuppositions and categorize them in Table~\ref{tab:data-analysis}. The five most frequent types of presuppositions include: %

\vspace{-.2em}
\begin{itemize}[leftmargin=1.3em]
    \setlength\itemsep{-0.1em}
    
    \item \textbf{False clauses} are those where \FP\ is made as a clause in \Q, e.g., ``the water has to be 100 to become steam'' or ``the motion of the fan generates heat'' in Table~\ref{tab:data-analysis}.
    
    \item \textbf{False predicate} are those where the predicate in \Q\ is false, e.g., ``current is stored in power plants'' or ``light bounce off objects''. This is the most common type of false presuppositions on \dataname.
    
    \item \textbf{False properties} are those where certain properties or attributes are presupposed in \Q, e.g., ``cement blocks are too strong so that people who punch them are likely to break their hand.''
    They are often very implicit and may be deniable.
    However, the question writer would not have asked the question otherwise, thus it is reasonable to assume that the \FP\ is actually made by the question writer.
    
    \item \textbf{False (causal) relationship between facts} are those where \Q\ makes a (causal) relationship between facts that are false. This is another very implicit type of presuppositions, but again, the question writer would not have asked this question if they have not made such a presupposition. An interesting observation is that a large portion of such questions involve scientific phenomenon
    on which the question writer has misunderstanding, e.g., in the example in Table~\ref{tab:data-analysis}, the question writer had a misconception about Newton's Third Law of Motion.
    
    \item \textbf{False existential presupposition} indicates that \Q\ includes an existential presupposition, one type of semantic presuppositions, that is false. For instance, the example \Q\ in Table~\ref{tab:data-analysis} presupposes an unused space in a hard disk, and \C\ says there is no unused space.
\end{itemize}

\paragraph{Triggers in the comment.}
We analyze how comments point out the falsehood of the presupposition made in the question on the same set of 50 samples on the development data.
In 68\% of times, comments include specific lexical cues which we call {\em triggers}.
70\% of such triggers are negations. Other word-level triggers include ``actually'', ``just'', ``rare'', ``really'' and ``though''. Sentence-level triggers include ``You are mistaken'', ``You've got some major confusion here'', ``I don't think you are fully comprehending ...'' and ``It should be noted that...''. 80\% of triggers appear in the first sentence of the comment, and the rest of the sentences elaborate on how the presupposition is false or provide other relevant information that does not directly answer the question.
The rest 32\% do not include lexical triggers and requires more careful comprehension of the comment with respect to the question, e.g., the false existential example in Table~\ref{tab:data-analysis}.

\vspace{-.3em}
\paragraph{Analysis of ambiguous cases.}
Even with our best efforts, there are still inherent disagreement between annotators.
Some of them are due to inherent ambiguities in language, e.g., the first example in Table~\ref{tab:ambiguity-analysis} where `the state of the water' could either mean the molecule itself or the energy state of the molecule.
Others are due to disagreement on the validity of the presupposition, e.g., in the second example in Table~\ref{tab:ambiguity-analysis}, it is debatable whether or not the question writer presupposes that the Board deals with day to day at a company.
We revisit this issue in human performance estimation in Section~\ref{subsec:detection-subtask-results}.

\begin{table}[!t]
    \centering \footnotesize
    \begin{tabular}{p{0.46\textwidth}}
        \toprule
            \multicolumn{1}{c}{\textbf{Ambiguity in language}} \\
            \Q\ When water boils, its bubbles are round, but when it freezes, its crystals are 6-sided. Why isn’t frozen water round or boiling water hexagonally shaped? Aren’t H2O molecules the same in either state? \\
            \C\ Bubbles are round because (...) Ice crystals are shaped in such a way because (...) The water molecules are much slower and aren't bouncing all over the place. Gaseous H2O is much higher energy and further apart so that the regular pattern of ice doesn't come into effect. \\
        \midrule
            \multicolumn{1}{c}{\textbf{Ambiguity in whether the presupposition is made}} \\
            \Q\ How do executives hold board of director positions at multiple fortune 500 companies? \\
            \C\ The Board only meets occasionally to vote on more important matters. They don't really deal with day to day affairs like the CEO does. \\
        \bottomrule
    \end{tabular}\vspace{-.1em}
    \caption{Two types of ambiguous cases. \Q\ and \C\ indicate the question and the comment, respectively.
    }\label{tab:ambiguity-analysis}
\end{table}

%% file: sections/04-task-setup.tex
The model is given a question $q$, and is required to perform the following subtasks:
\vspace{-.3em}
\begin{itemize}\itemsep -.1em
    \item[(a)] Detection subtask: assign a label to be  \fp\ or \n; \fp\ means $q$ has a false presupposition, and \n\ means $q$ has no false presuppositions.
    We use a macro-F1 score as an evaluation metric. 
    \item[(b)] Writing subtask: if the label is \fp, write the false presupposition as well as the correction.
    We use  sacreBLEU~\citep{post-2018-call} and unigram-F1 following \citet{petroni-etal-2021-kilt}. %
    We also introduce a human evaluation scheme in Section~\ref{subsec:writing-subtask-human}.
\end{itemize}
We have two tracks: the main track and the \NLITrack.

\vspace{.3em}
\noindent \textbf{The main track} provides $q$ as the only input to the model. The model is expected to search necessary background knowledge to perform the task from any information source except for Reddit and Quora.\footnote{
This is because similar questions are being asked on these websites. The same decision has made in \citet{nakano2021webgpt}.
}
This is the most realistic setting for the typical open-domain question answering problem.

\vspace{.3em}
\noindent \textbf{The \NLITrack} provides the comment used for the annotation as an additional input to the model. This removes the need for retrieval, and guarantees that all necessary information to perform the task is given.

%% file: sections/05-exp-detection.tex
This section discusses baseline experiments for the detection subtask; Section~\ref{sec:exp-writing} discusses baseline experiments for the writing subtask.

\subsection{Baselines}\label{subsec:detection-subtask-baselines}

\subsubsection{Trivial baselines}\label{subsec:trivial-baselines}
\noindent
\textbf{Random} assigns \fp\ or \n\ randomly at uniform.
\textbf{\fp\ only} always assigns \fp. 
\textbf{\n\ only} always assigns \n. 
\textbf{Nearest Neighbor} retrieves one of questions from the training set that is closest to the test question, based on c-REALM~\citep{krishna-etal-2021-hurdles}, and returns its label as the prediction.

\subsubsection{\NLITrack\ baselines}\label{subsec:oracle-comment-baselines}
\textbf{Question only} trains a RoBERTa-based~\citep{liu2019roberta} classifier that takes the question as the only input and classifies the label. It is often called closed-book model~\citep{roberts-etal-2020-much}.
\textbf{Comment only} is a classifier based on RoBERTa-large that takes the comment as the only input and assigns the label; this baseline is provided as a reference.
\textbf{Question$\oplus$Comment} is a classifier based on RoBERTa-large that takes a concatenation of the question and the comment to the classifier, and assigns the label. %
We additionally experiment with the same model that is trained on either MNLI~\citep{N18-1101} or BoolQ~\citep{clark-etal-2019-boolq}, and tested on \dataname\ in a zero-shot fashion. This condition tests if training on similar, previously studied datasets helps.

\subsubsection{Main track baselines}\label{subsec:main-baselines}
We design a model called \textbf{c-REALM + MP (Multi-passage) classifier} that retrieves a set of paragraphs from Wikipedia and then assigns a label.

First, the model uses c-REALM~\citep{krishna-etal-2021-hurdles}, a state-of-the-art retrieval model on ELI5, to retrieve a set of $k$ passages from the English Wikipedia.
Next, the model uses the multi-passage classifier based on RoBERTa in order to assign a label. Given a question $q$ and a set of passages $p_1...p_k$, each $p_i$ ($1\leq i \leq k$) is concatenated with $q$ and is transformed into $\mathbf{h}_i \in \mathbb{R}^{h}$ through the Transformer model.
We then obtain logits via $\mathbf{p} = \mathrm{FFN}(\mathrm{MaxPool}(\mathbf{h}_1...\mathbf{h}_k)) \in \mathbb{R}^2$, where $\mathrm{FFN}$ is a feed-forward layer and $\mathrm{MaxPool}$ is an element-wise max operator.
Finally, we use $\mathrm{Softmax}(\mathbf{p})$ to compute the likelihood of $q$ having false presuppositions or not.

\vspace{-.3em}
\paragraph{Self-labeling}
Although our labeled data is small, there is large-scale unlabeled data (question and comment pairs) available.
We explore self-labeling to leverage this unlabeled data.
Specifically, we use the \textbf{Question$\oplus$Comment} to assign a silver label to the unlabeled training questions.
We then train the classifier on the union of this silver data as well as the gold labeled data.

\subsubsection{Human performance}
We estimate human performance to better understand the model performance. We recruit two human workers who perform the task for 186 questions for each track.

We estimate two types of human performance.
(1) \textbf{Human with the most voted comment}, where human workers assume the most voted comment as a ground truth in terms of factuality of the information and the validity of the presupposition. We think of it as an upperbound of model performance, since the comment is from multiple online users or domain experts.
(2) \textbf{Human w/o the most voted comment}, where human workers search over the web (except Quora and Reddit) to find information, and make the best judgment about the validity of the presupposition. We think it is likely to be worse than the upperbound of model performance, since only one worker, instead of multiple online users or domain experts, makes a decision.

\subsection{Results}\label{subsec:detection-subtask-results}

Results are reported in Table~\ref{tab:results-detection-task}.

\vspace{-.3em}
\paragraph{The \NLITrack.}
First, all trivial baselines achieve poor performance. In particular, poor performance of the nearest neighbor model indicates that there is no significant train-test overlap on \dataname. 
The question only baseline and the comment only baseline are more competitive. This is likely due to bias in the question distribution (in case of the question only), and the comment usually includes information from the question (in case of the comment only).
Nonetheless, using both the question and the comment (Question$\oplus$Comment) achieves the best performance, outperforming the best trivial baseline by 22\% absolute.
Zero-shot models trained on MNLI and BoolQ achieve poor performance, being even worse than some of trivial baselines. This indicates our problem is significantly different from existing tasks like NLI or binary question answering.
The best model is 10\% below human performance, indicating room for improvement, even in the easier track.

\begin{table}[!t]
    \centering \footnotesize
    \begin{tabular}{lrr}
        \toprule
            Model & Dev & Test \\
        \midrule
            \multicolumn{3}{l}{\textbf{\em{Trivial baselines}}} \\
            Random$^\otimes$  &  44.9  & 47.8  \\
            Always predict \fp$^\otimes$  & 21.4 & 20.0 \\
            Always predict \n$^\otimes$ & 42.1 & 42.9   \\
            Nearest Neighbor$^\otimes$ & 56.2 & 54.1 \\
        \midrule
            \multicolumn{3}{l}{\textbf{\em{\NLITrack}}} \\
            Question only & 67.7 & 66.9  \\
            Comment only & 68.9 & 68.6 \\
            Question$\oplus$Comment & \textbf{76.3} & \textbf{75.6}   \\
            Question$\oplus$Comment (MNLI)$^\otimes$ & 54.4 &  54.2 \\
            Question$\oplus$Comment (BoolQ)$^\otimes$ & 60.4 &  58.2\\
        \midrule
            \multicolumn{3}{l}{\textbf{\em{Main track}}} \\
            c-REALM + MP classifier & 68.3 & 66.3 \\
            c-REALM + MP classifier (Self-labeling)$^\ddagger$ & \textbf{69.1} & \textbf{67.1} \\
        \midrule
            Human w/ most-voted comment & 86.4 & 85.1 \\
            Human w/o most-voted comment & 70.9 & 70.9 \\
        \bottomrule
    \end{tabular}\vspace{-.1em}
    \caption{Baseline results in the \textbf{detection subtask} on the development data and the test data, respectively. \textbf{Macro-F1} scores reported.
    By default, the models are trained on the labeled portion of \dataname;
    $^\otimes$ indicates the model is not trained on \dataname; $^\ddagger$ indicates the model is trained on both the labeled and unlabeled portions of \dataname.
    }
    \label{tab:results-detection-task}
\end{table}

\vspace{-.3em}
\paragraph{The main track.}
Using retrieved passages from c-REALM and multi-passage classifier achieves 66.3\% 
on the test set, which is significantly better than all trivial baselines.
The self-labeling technique leads to additional improvements, leading to an F1 of 67.1\%.
While these numbers are significantly better than trivial baselines, they are significantly worse than the model performance given the comment, and in fact, is very close to the performance of the question only baseline. This strongly suggests a retrieval bottleneck---getting passages that provide evidence as strong as human-written comments is difficult even with the state-of-the-art retrieval model.

To further support the bottleneck in retrieval, we conduct a detailed error analysis in Appendix~\ref{app:error-analysis}. For instance, 86\% of false negatives were due to retrieval misses, including failing to retrieve relevant topics (42\%), retrieving evidence on the relevant topics but not related to the false presuppositions (32\%), or retrieving evidence related to the presuppositions but is not direct enough (12\%).\footnote{This can be seen as a classification error---if the classification model can better capture implicit evidence, it could have made a correct prediction.}

\paragraph{Human performance.}
Humans given the most upvoted comment achieve performance that is significantly higher than all baseline numbers, indicating significant room for improvement.

Without the most upvoted comment, people achieve relatively poor performance
(70.9\%).
To better understand this, we analyze 44 error cases, and categorize them in Table \ref{tab:human-perf-detection-task}.
Nearly half of the errors are due to an inherent disagreement in labels, either due to (1) ambiguity, either in language or whether the presupposition was made, or (2) whether it is critical to correct false presuppositions (especially cases in the exception category in Table~\ref{tab:data-analysis}).
The most upvoted comment is an aggregation of many active users of the community, thus it is reasonable to consider the most upvoted comment as a ground truth, and human performance with the most upvoted comment as an upperbound performance of \dataname.
Future work may take other approaches, e.g., not treating it as a binary classification problem.

\begin{table}[!t]
    \centering \footnotesize
    \begin{tabular}{p{0.35\textwidth}r}
    \toprule
        Error Category &\% \\ \midrule
        Failure in finding evidence &  11.4 \\
        Mistakes in labeling & 11.4 \\
        Wrong ground truth label &  11.4 \\
        Inherent disagreement: ambiguity & 34.1 \\
        Inherent disagreement: criticalness
        & 9.1 \\
        Information on the web being inconsistent & 22.7  \\
    \bottomrule
    \end{tabular}\vspace{-.1em}
    \caption{Analysis of 44 errors made by human performers without the most upvoted comment.
    }
    \label{tab:human-perf-detection-task}
\end{table}

\newcommand{\result}[3]{#2 & #3 & #1}

\begin{table*}[!t]
    \centering \footnotesize
    \begin{tabular}{l
        R{.7cm}R{.7cm}R{.7cm}R{.7cm}
        R{.7cm}R{.7cm}R{.7cm}R{.7cm}
        R{.7cm}R{.7cm}R{.7cm}R{.7cm}
        }
        \toprule
            \multirow{3}{*}{Model} &
            \multicolumn{6}{c}{Development} & \multicolumn{6}{c}{Test} \\
            & \multicolumn{3}{c}{uF1} & \multicolumn{3}{c}{BLEU}
            & \multicolumn{3}{c}{uF1} & \multicolumn{3}{c}{BLEU}
            \\
            \cmidrule(lr){2-4} \cmidrule(lr){5-7} \cmidrule(lr){8-10} \cmidrule(lr){11-13}
            & P & C & A & P & C & A & P & C & A & P & C & A  \\
        \midrule
            \multicolumn{5}{l}{\textbf{\em{\NLITrack}}: Copy Baseline} \vspace{.3em} \\
            & \result{35.8}{45.0}{26.5}
            & \result{10.2}{14.8}{5.5}
            & \result{35.9}{44.7}{27.0}
            & \result{10.2}{14.6}{5.7}
            \\
        \midrule
            \multicolumn{5}{l}{\textbf{\em{\NLITrack}}: Question$\oplus$Comment} \\
            ~~Dedicated
            & \result{\textbf{44.8}}{51.7}{\textbf{37.8}}
            & \result{\textbf{24.0}}{30.0}{\textbf{17.9}}
            & \result{\textbf{42.9}}{47.7}{\textbf{38.0}}
            & \result{\textbf{19.5}}{22.9}{\textbf{16.1}}
            \\
            ~~Unified
            & \result{43.3}{\textbf{53.4}}{33.2}
            & \result{21.5}{\textbf{30.9}}{12.0}
            & \result{40.3}{\textbf{49.2}}{31.4}
            & \result{18.2}{\textbf{25.9}}{10.4}
            \\
        \midrule
            \multicolumn{5}{l}{\textbf{\em{Main track}}: Question + c-REALM} \\
            ~~Dedicated
            & \result{37.9}{49.1}{26.6}
            & \result{18.5}{28.2}{8.7}
            & \result{35.4}{45.9}{24.9}
            & \result{14.9}{22.6}{7.1}
            \\
            ~~Unified
            & \result{40.1}{49.4}{30.7}
            & \result{19.5}{28.8}{10.1}
            & \result{37.4}{46.3}{28.4}
            & \result{16.0}{23.6}{8.3}
            \\
        \bottomrule
    \end{tabular}\vspace{-.1em}
    \caption{Baseline results in the \textbf{writing subtask}, on the development data and the test data, respectively. We report unigram F1 (uF1) and BLEU.
    P, C and A indicate Presupposition, Correction, and Average between two.
    }\label{tab:results-writing-task}
\end{table*}

%% file: sections/06-exp-writing.tex
\subsection{Baselines}\label{subsec:writing-subtask-baselines}
For the writing subtask, the system is given a question that is guaranteed to contain a false presupposition, and is required to generate the presupposition as well as the correction.

\subsubsection{\NLITrack\ baselines}\label{subsec:gold-comment-writing-baselines}

\paragraph{Copy baseline.} As a trivial baseline, we copy the given question as a presupposition and the given comment as a correction.

\vspace{-.3em}
\paragraph{Question$\oplus$Comment Dedicated.}
We train two generators separately to generate the presupposition and the correction, respectively, given a concatenation of the question and the comment.
Both models are based on the pretrained T5-base model ~\citep{2020t5}.

\vspace{-.3em}
\paragraph{Question$\oplus$Comment Unified.}
While having two dedicated models is a straight-forward way to produce two writings, we design a unified model that can be used for both the presupposition and the correction, motivated by the intuition that generation of each can benefit from each other.

We train one generator that is essentially to generate the correction. Specifically, the model is trained with a union of (1) annotated corrections, and (2) annotated presuppositions
prepended with ``It is not the case that'' so that they look like corrections.
At inference time, we use a standard, beam search decoding to generate the correction. To generate the presupposition, we first decode a sequence with a constraint~\citep{decao2020autoregressive} that it should start with ``It is not the case that'', and then take the sequence that comes next as a presupposition.

\subsubsection{Main track baselines}\label{subsec:main-writing-baselines}

We design \textbf{c-REALM + MP (Multi-Passage) Dedicated} and \textbf{c-REALM + MP (Multi-Passage) Unified}. They are similar to the dedicated and unidifed models in Section~\ref{subsec:gold-comment-writing-baselines}.
The only difference is that the model receives a question and a set of $k$ passages from c-REALM instead of a question-comment pair.
In order for the T5 model to read multiple passages,
we use the Fusion-in-Decoder architecture~\citep{izacard-grave-2021-leveraging}. We refer to Appendix~\ref{app:model-details} for more details.

\begin{table}[!t]
    \centering \footnotesize
    \begin{tabular}{lR{0.7cm}R{0.7cm}R{0.7cm}R{0.7cm}}
        \toprule
            Model & F & P & \textsc{Cr} & \textsc{Cs} \\
        \midrule
            \multicolumn{5}{l}{\textbf{\em{\NLITrack}}: Question + Comment} \\
            ~~Dedicated & 2.9 & 1.8 & 1.9 & 1.6 \\
            ~~Unified & 3.0 & 2.0 & 0.8 & 2.8 \\
        \midrule
        \multicolumn{5}{l}{\textbf{\em{Main Track}}: Question + c-REALM} \\
            ~~Dedicated & 2.8 & 1.8 & 0.6 & 1.6 \\
            ~~Unified & 3.0 & 1.8 & 0.6 & 2.8 \\
        \midrule
        \textbf{\em{Groundtruth}} & 2.9 & 2.8 & 2.7 & 2.9 \\
        \midrule
        \textbf{Agreement (\%)} & 96.3 & 65.4 & 63.0 & 74.8 \\
        \bottomrule
    \end{tabular}\vspace{-.1em}
    \caption{Human evaluation results (scale: 0--3). F: Fluency, P: Presupposition, \textsc{Cr}: Correction, \textsc{Cs}: Consistency.
    The last row reports the inter-annotator agreement rate.
    }\label{tab:human-eval-results-writing-task}
\end{table}

\subsection{Results: Automatic Evaluation}\label{subsec:writing-subtask-results}

Table~\ref{tab:results-writing-task} reports the results. Examples of model outputs are provided in Appendix~\ref{app:writing-subtask-details}.

For \NLITrack, the dedicated model gives the best result across all metrics except unigram F1, outperforming the copy baseline and the unified model.
On the other hand, for the main track, the unified model outperforms the dedicated model.
We think this is because when the comment is given, finding the presupposition from the question and the correction from the comment in isolation is sufficient; however, in the main track where the comment is not given, the model benefits from multi-task learning.

For both tracks, models are better in writing presuppositions than writing corrections,
likely because the presupposition can often be extracted from the question alone, while the correction needs more comprehensive understanding of both the question and the comment, or an extra information from retrieval.
Although performance on presupposition identification is similar in the \NLITrack\ and in the main track, correction performance is significantly higher in the \NLITrack. We think it is because retrieving evidence passages that contain correction of the false presupposition is challenging, as discussed in Section~\ref{subsec:detection-subtask-results}.

\subsection{Results: Human Evaluation}\label{subsec:writing-subtask-human}
To augment the automatic metrics, we conduct human evaluation of model generations of 200 randomly sampled test instances from 4 aspects: fluency, presupposition, correction and consistency. 
\vspace{-.8em}
\begin{itemize}[leftmargin=1.3em]
    \setlength\itemsep{-0.1em}
    \item \textbf{Fluency}: The generated text should be fluent (i.e., free of grammatical errors, spelling errors, and repetitions).
    \item \textbf{Presupposition}: The generated presupposition should be the valid one in the question, and is factually false. 
    \item \textbf{Correction}: The correction should be made in the comment and provide reasonable amount of justification rather than being a simple negation of the presupposition.
    \item \textbf{Consistency}: The presupposition and correction should be on the same topic and %
    negate each other. 
\end{itemize}
Each aspect is rated in the 0--3 scale.
We evaluate the output from all systems except the copying baseline, as well as the ground truth reference.
Each question is assigned two raters in order to reduce noise and report inter-rater agreement on pairwise comparison.
More details about the rating scheme are provided in Appendix~\ref{app:writing-subtask-details}. 

Table~\ref{tab:human-eval-results-writing-task} reports the result of human evaluation.
All model can generate almost flawless fluent text, and generate presuppositions that are valid ones.
However, their outputs generated as false presuppositions are factually correct in half of the cases.
These observations are relatively consistent across different systems.

Notable differences between systems are found in correction and consistency. The dedicated model generates better correction, likely because it is given a comment. All other models struggle: in particular, the unified models tend to generate the correction that starts with ``It is not the case that'' even the model is not restricted to do so at inference time.
On the other hand, the unified model is better in consistency, likely because the dedicated model is more vulnerable in generating the presupposition and the correction in a totally different topic.

%% file: sections/07-discuss.tex
\paragraph{False Presuppositions in Other Data.}
While we focus on questions from an online forum due to the availability of large unlabeled data and the domain being fairly general, we argue that false presuppositions are not specific to such domains.
In fact, false presuppositions are more prevalent when the domain is specific and requires expertise.

We analyze 50 random samples of unanswerable questions from QASPER~\citep{dasigi-etal-2021-dataset}, a dataset consisting of information-seeking questions on NLP research papers, posed by NLP experts. 
We find that, out of 48 questions that are unanswerable (2 of them turn out to have valid answers), 25\% of them has false presuppositions, because the question writer does not have sufficient background knowledge or misses facts in the research paper.
Table~\ref{tab:other-data} shows a subset of such questions, along with false presuppositions and their correction annotated by the author.
Identification of false presuppositions requires external knowledge beyond the specific paper the question writer is reading (\myhighlightblue{1}), or requires understanding of details of the paper which may not be explicitly written with the exact same terms in the paper (\myhighlightblue{2} and \myhighlightblue{3}).

It should be noted that this percentage is a strict lower bound of true percentage and may be significantly underestimated since identification of false presupposition requires knowledge in NLP; the estimate will be higher when annotated by multiple NLP experts.
Moreover, presuppositions will significantly more prevalent when question writers are non-expertise, unlike in QASPER whose question writers are NLP experts.

We did not annotate and experiment with QASPER because the data is relatively small (272 and 62 unanswerable questions on the training set and on the validation set, respectively), but future work can investigate this data as well we domain transfer of models.

\begin{table}[!t]
    \centering \footnotesize
    \begin{tabular}{p{0.45\textwidth}}
        \toprule
            \myhighlightblue{1} \textbf{\em{Paper title:}} Combating Adversarial Misspellings with Robust Word Recognition \\
            \textbf{\em{Q:}} Why do they experiment with RNNs instead of transformers for this task? \\
            \textbf{\em{FP:}} The paper does not experiment with transformers. \\
            \textbf{\em{Corr:}} The paper uses RNNs and BERT. The question writer either missed the fact that they used BERT, or did not know that BERT is based on transformers. \\
        \midrule
            \myhighlightblue{2} \textbf{\em{Paper title:}} Analysis of Wikipedia-based Corpora for Question Answering \\
            \textbf{\em{Q:}} Can their indexing-based method be applied to create other QA datasets in other domains, and not just Wikipedia? \\
            \textbf{\em{FP:}} Their indexing-based method is applied to create a QA dataset in the Wikipedia domain. \\
            \textbf{\em{Corr:}} Their indexing-based method is not for creating QA datasets. This is for aligning (already annotated) answer context to a particular Wikipedia corpus. \\
        \midrule
            \myhighlightblue{3} \textbf{\em{Paper title:}} Automatic Classification of Pathology Reports using TF-IDF Features \\
            \textbf{\em{Q:}} How many annotators participated? \\
            \textbf{\em{FP:}} There are annotators. \\
            \textbf{\em{Corr:}} There is no annotators. 
            The paper created a dataset, but the data construction process is entirely automatic. \\
        \bottomrule
    \end{tabular}\vspace{-.1em}
    \caption{Example questions with false presuppositions on QASPER~\citep{dasigi-etal-2021-dataset}.
    {\em Q}, {\em FP} and {\em Corr} indicate the question, false presupposition, and the correction, respectively.
    }\label{tab:other-data}
\end{table}

\vspace{-.3em}
\paragraph{A case study with GPT-3.}
Large language models such as GPT-3~\citep{NEURIPS2020_1457c0d6} have shown impressive performance in generating a response to the question. We conduct small-scale evaluation of Instruct GPT-3 (\texttt{text-davinci-002}), whose details are not public but is known as the best version of GPT-3.
An example is depicted in Table~\ref{tab:gpt-3}.
We find that most generations are roughly on the right topic, e.g., all generations in Table~\ref{tab:gpt-3} discuss Newton's Law of Motion.
However, they rarely correctly satisfy users information need:
\vspace{-.2em}
\begin{itemize}[leftmargin=1.3em]
    \setlength\itemsep{-0.1em}
    \item Most of them include information that is factually false, e.g., the equal and the opposite action are not instantaneous (\myhighlight{1} and \myhighlight{4}), their magnitude is not equal (\myhighlight{2} and \myhighlight{6}), or they do balance out (\myhighlight{3}).
    \item They are often not precisely about what the question is asking about. For instance, they discuss why an object may not move given a force, e.g., the wall does not move when you hit the wall (\myhighlight{2} and \myhighlight{6}). This is related to Newton's Law of Motion, but not at all to the question.
    \item They do not explicitly identify false presuppositions. None of the generation mentions that the key misunderstanding is that the equal and opposite reaction apply to different objects, thus are not cancelled out. Sometimes the generation indicate some part of the question is wrong (indicates with `but' or `However') but does not precisely point out what is wrong, nor provide corrections.
\end{itemize}

\vspace{-.3em}
It is possible performance could be improved with better prompting, but we leave this possibility to future work.

\begin{table*}[!t]
    \centering \footnotesize
    \begin{tabular}{p{0.98\textwidth}}
        \toprule
            \textbf{\em{Q:}} If there's an equal and opposite reaction for everything, how does any action happen? Isn't it balanced out by the opposite reaction? \\
            \textbf{\em{FP:}} The equal and opposite reaction apply to the same object. \\
            \textbf{\em{Corr:}} Based on Newton's Law of Motion, the equal and opposite reaction apply to different objects. Only forces that are applied to the same object would be cancelled out. \\
        \midrule
            \myhighlight{1} \textbf{GPT-3 given Q, temp=0.0}: The equal and opposite reaction is not always instantaneous, so an action can still happen even if there is an equal and opposite reaction. \\
            ~~~\mycomment\ Factually false. The equal and opposite reaction are always instantaneous. \\
            \myhighlight{2} \textbf{GPT-3 given Q, temp=0.7}: According to Newton's third law of motion, ``For every action, there is an equal and opposite reaction.'' However, this only applies to objects interacting with other objects. In other words, if Object A exerts a force on Object B, then Object B will exert an equal and opposite force on Object A. However, if there is no other object for the force to be exerted on, then there is no equal and opposite reaction, and the force will simply cause the object to accelerate in the direction of the force. \\
            ~~~\mycomment\ Does not explicitly point out \fp, and the correction of \fp\ is factually wrong.  \\
            \myhighlight{3} \textbf{GPT-3 given Q, temp=0.7 (different sample)}: There is an equal and opposite reaction for every action, but this doesn't mean that every action is balanced out by the opposite reaction. For example, if you push on a wall, the wall pushes back on you with an equal and opposite force.
            However, the wall doesn't move because the force you're exerting on it is balanced out by the force of the wall pushing
            back on you. \\
            ~~~\mycomment\ The correction of \fp\ is factually wrong. In the scenario mentioned in the generation, the wall does not move because its mass is large.
            Besides, this scenario is not related to what the question is asking about. \\
            \myhighlight{4} \textbf{GPT-3 given Q + ``Or did I miss anything?'', temp=0.7}:
            No, you didn't miss anything. The equal and opposite reactions do balance each other out, but the net result is still an
            action. This is because the reactions don't happen instantaneously - there is a brief period of time during which one
            reaction is occurring without the other. During this time, the system is unbalanced and an action can occur. \\
            ~~~\mycomment\ Factually false, and fail to identify \fp. The equal and opposite reactions do not balance out, and they happen instantaneously. \\
           \myhighlight{5}  \textbf{GPT-3 given Q + ``Or did I misunderstand anything?'', temp=0.7}:
            Newton's third law of motion states that for every action, there is an equal and opposite reaction. However, this does
            not mean that every action is perfectly balanced by an equal and opposite reaction. Instead, it means that the total
            force exerted by all objects on each other is always zero. \\
            ~~~\mycomment\ Does not explicitly point out \fp, and provide adequate correction. \\
            \myhighlight{6} \textbf{GPT-3 given Q + ``Or do I have any misunderstanding?'', temp=0.7}:
            There is an equal and opposite reaction for every action, but not every reaction is equal in magnitude to the action. For
            example, if you push on a wall, the wall pushes back on you with an equal and opposite force, but the force exerted by
            the wall is much less than the force you exerted. This is because the wall is much more massive than you are, so it
            takes more force to move it. \\
            ~~~\mycomment\ Factually false.
            The force exerted by `you' and the force exerted by the wall is equal. Besides, this is not what the question is asking about. It is answering a different question: why the wall won't move given a force from you. \\
        \bottomrule
    \end{tabular}\vspace{-.1em}
    \caption{Response from Instruct GPT-3 (\texttt{text-davinci-002}) given an example question.
    temp=0.0 is deterministic, and temp=0.7 is the default value provided by GPT-3.
    Generations are obtained by sampling multiple times, or using a slightly different templates. We did not cherry-pick these generations: we ran the model six times and report all.
    }\label{tab:gpt-3}
\end{table*}

%% file: sections/08-concl.tex
In this paper, we identified that false presuppositions are prevalent in information-seeking user questions.
We introduced \dataname: the first benchmark for the correction of false presuppositions in the open-domain setup.
\dataname\ consists of 8,400 user questions, 25\% of which contain false presuppositions and are paired with their corrections.
Our detailed analysis highlights challenges in solving the task, including (1) retrieval of evidence that identifies false presupposition, (2) identification of implicit and subtle presuppositions, and (3) generating correction that is accurate and adequately explains how and why the presupposition is false.
We hope our benchmark adds to the problem of open question answering, inviting researchers to build models to study questions with false presuppositions. Further, we suggest future work to develop better models, explore approaches to address inherent debatability of the judgment, and evaluation of the model generation.

%% file: sections/10-app.tex
\section{Details in Data Source}\label{app:data-source-details}

\paragraph{Filtering Data Sources.}
When the question has several comments, we choose the comment with the highest upvotes.
To remove toxic language on the data, we follow the toxicity word list from \citet{cachola-etal-2018-expressively} to remove questions that contain any of the toxic words, except ``hell'' and ``damn'', as these two words are commonly used as interjections. %

\paragraph{Analysis in Train-Test Overlap.}
\citet{krishna-etal-2021-hurdles} reported that 81\% of the validation questions of the original ELI5 dataset are paraphrases of the question on the training set. We revisit this issue, and show with a careful assessment of underlying assumptions and a finer-grained definition of ``paraphrases'', the proportion of paraphrased questions is significantly smaller.

We assess 104 randomly sampled questions in the validation set with their closest 7 training questions retrieved using c-REALM, as \citet{krishna-etal-2021-hurdles} did, but with a different rating scale (1--4), following \citet{bhagat-hovy-2013-squibs}, \citet{ganitkevitch-etal-2013-ppdb}, and \citet{pavlick-etal-2015-ppdb}:

\vspace{-.2em}
\begin{itemize}[leftmargin=1.3em]
    \setlength\itemsep{-0.1em}
    \item \textbf{1: No Paraphrase; No similar intention}: Two questions do not share the meaning nor the intention.
    \item \textbf{2: No Paraphrase; Similar intention}: Two questions are likely to have the same intention, but they are not paraphrases, because their literal meanings are different and/or their underlying assumptions are different.
    \item \textbf{3: Non-trivial paraphrase:} 
        Most of the two questions' meanings are the same; however, they do not belong to any of lexical paraphrase (single word to single word), phrasal paraphrase (multiword to single/multiword), or syntactic paraphrase (paraphrase rules containing non-terminal symbols),\footnote{Definition derived from \citet{ganitkevitch-etal-2013-ppdb}.} and require non-trivial background knowledge to identify whether they have the same meaning.
    \item \textbf{4: Paraphrases}: Two questions fall into either lexical paraphrase, phrasal paraphrase, syntactic paraphrase, structured paraphrase, or other trivial paraphrase.
\end{itemize}

Table~\ref{tab:paraphrase} presents the percentage and an example of each category. 76.9\% questions have a rating of 2 or less, indicating that for most validation question, there are either no similar intention question in the training set, or there are questions with similar intention but either their literal meanings are different or their underlying assumptions are different. In particular, the latter indicates that whether or not there is a false presupposition can be different, even though they may share similar intention.
Only 23.1\% questions have a rating of 3 and above, indicating that relatively few questions in the validation set have a non-trivial or trivial paraphrase in the training set.

We also explored automatically filtering paraphrased questions using BLEU or TF-IDF. However, we find it is non-trivial to find the right threshold, thus include all questions and leave filtering to future work.

\begin{table}[!t]
    \centering \footnotesize
    \begin{tabular}{lp{0.3\textwidth}}
        \toprule
        \textbf{1} (45.2\%) & \textbf{\em{Dev}}:  Why are aluminium alloys difficult to weld? \\
        &\textbf{\em{Train}}: Cold welding. Two pieces of metal touch in a vacuum, why do they stick together? How strong is this weld? \\
        \midrule 
        \textbf{2} (31.7\%) & \textbf{\em{Dev}}: How is blood after a transfusion integrated into the body, especially when the transfused RBCs carry different DNA than the host RBCs? \\
        &\textbf{\em{Train}}: How DNA from blood is changed when getting a blood transfusion\\
        &\mycomment\ %
        The dev question assumes that the transfused RBCs carry DNA, while the train question does not.
        \\
        \midrule
        \textbf{3} (6.7\%) & \textbf{\em{Dev}}: How is information retained in solid-state memory devices after power is turned off?\\
        &\textbf{\em{Train}}: How do electronics keep memory after you take all power sources away?\\
        & \mycomment\ It is not trivial that ``information retained in solid-state memory devices'' is a paraphrase with ``electronics keep memory''. \\
        \midrule
        \textbf{4} (16.3\%) &\textbf{\em{Dev}}: What is the difference between centrifugal and centripetal force? \\
        &\textbf{\em{Train}}: The difference between centrifugal force and centripetal force.\\
        \bottomrule
    \end{tabular}
    \caption{The rating scale for paraphrase and examples for each category.
    }\label{tab:paraphrase}
\end{table}

\section{Details in Data Annotation}\label{app:data-annotation-details}

\begin{figure*}[t]
\includegraphics[width=\textwidth, keepaspectratio]{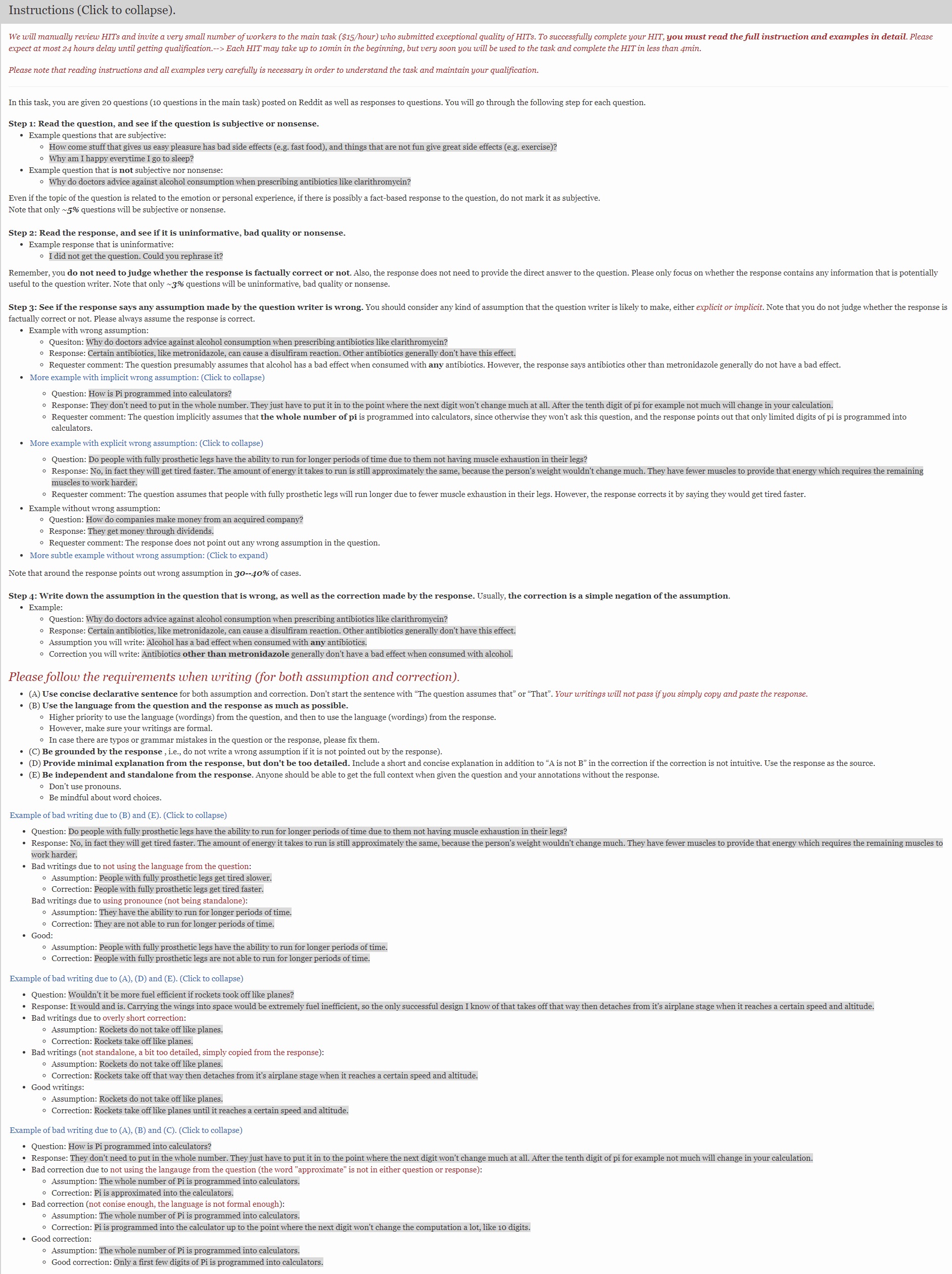}\vspace{-.1cm}
\caption{
    The instruction we provided for our qualification task.
}\label{fig:instruction}
\end{figure*}

\begin{figure*}[t]
\includegraphics[width=\textwidth, keepaspectratio]{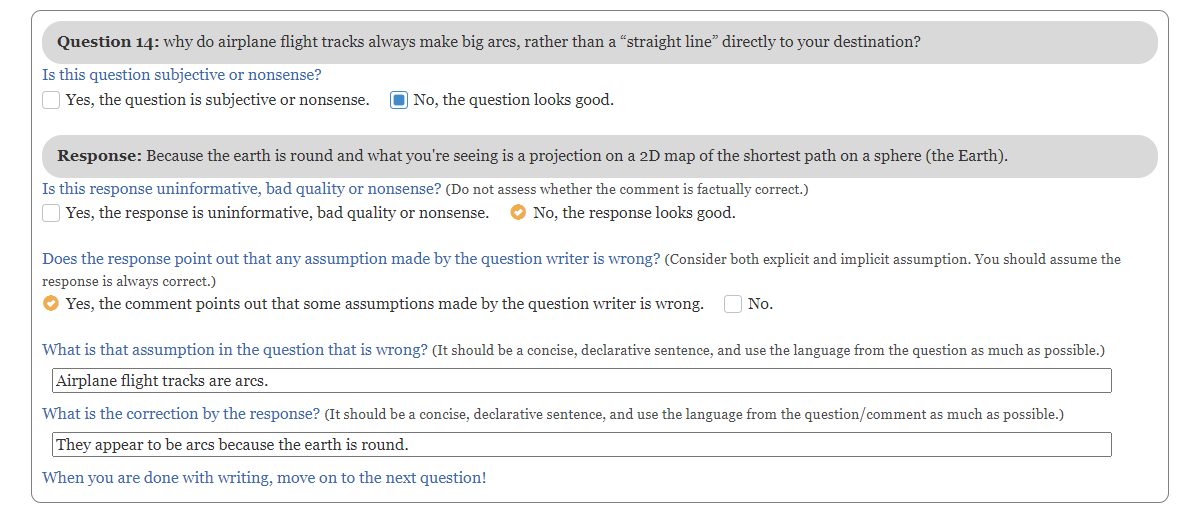}\vspace{-.1cm}
\caption{
    An example of our designed annotation interface.
}\label{fig:interface}
\end{figure*}

The annotation instruction is in Figure \ref{fig:instruction}, and we show an example of the annotation interface in Figure \ref{fig:interface}.

\paragraph{Qualification Task.} The qualification task contains 20 pre-annotated questions by the authors, and we provide examples as well as their explanation to workers to demonstrate our task. Based on both whether the worker can correctly identify false presupposition and the writing quality, we selected 30 qualified workers. 

\paragraph{Generation Task.} Qualified workers who passed the qualification task work on the main annotation task. Each question is assigned to two generators who independently annotate the label.
We monitor the agreement of workers and send the disagreed cases to further validate. We revoke qualification of generators whose more than 10\% of annotations marked to be invalid by the validators.

\paragraph{Validation Task.}
If two generators disagree on the label, their annotations are sent to two validators who judge their validity.
We select a smaller number of high qualified workers who have exceptional understanding of false presuppositions and are active users of Reddit.
We find that for a small number of highly ambiguous cases, two validators have disagreement. In this case, we send the question to a third validator and take the majority vote. 

\vspace{.4em}
Generators and validators are paid with reasonable hourly wage (13 USD/hour and 18 USD/hour, respectively).

\section{Details in Experiments}\label{app:exp-details}

\subsection{Model details}\label{app:model-details}
\paragraph{Retrieval model.}
We use the English Wikipedia from 08/01/2019 provided by \citet{petroni-etal-2021-kilt}.
We obtain c-REALM embeddings of Wikipedia passages as well as the questions.
We use FAISS~\citep{johnson2019billion} to do approximate maximum inner product search to retrieve the top-5 passages for the query. 

\paragraph{Classifier model.}
For all the models in the \NLITrack\ track, we use a per GPU batch size of 2, and for all the models in the main track, we use the top-5 passages as the context and a per GPU batch size of 8. We train all our model with learning rate $10^{-5}$, weight decay of 0, and maximum sequence length of 256. We train all our models for 5000 steps and a gradient accumulation step of 5.

All experiments were done with two Nvidia-RTX6000 GPUs (the detection subtask) or two Nvidia-A40 GPUs (the writing subtask). 
We use half-precision training~\citep{micikevicius2018mixed} when training for our detection task and gradient checkpointing from fairscale~\citep{FairScale2021}, and choose the best checkpoint based on the performance on the validation set and report its test set performance. For the unified models for the writing subtask, we choose the best checkpoint that gives the best BLEU score for presupposition.

\subsection{Error analysis on the detection subtask}
\label{app:error-analysis}

\begin{table}[!t]
    \centering \footnotesize
    \begin{tabular}{p{1.5cm}p{2.8cm}|rr}
    \toprule
        \multicolumn{2}{c}{Error Type} & FP(\%) & FN(\%) \\\midrule
         \textbf{Retrieval} &  No related topic & 26 & 32\\
         & Similar topic but not enough to make decision & 20 & 40\\
         & Indirect evidence and require reasoning & -- & 12\\\midrule
         \textbf{Classification} &  Direct evidence & 34 & 10\\\midrule
         \textbf{Labeling} & Ground truth label is wrong & 8 & 8 \\\midrule
         \textbf{Inherent Disagreement} & Ambiguity & 4 & 2 \\
         & Criticalness & 2 & 0\\\midrule
         \textbf{Inconsistency} & Information on the web being inconsistent & 10 & 4 \\
    \bottomrule
    \end{tabular}
    \caption{The breakdown analysis of false positive/false negative in the validation set for c-REALM + MP classifier model. FP: False positive. FN: False negative. \fp~: False Presupposition. ``Indirect evidence and require reasoning'' can also belong to retrieval error. Information on the web being inconsistent: the comment contradicts with the retrieved passages. The categories are \textit{not} mutually exclusive.
    }
    \label{tab:detection-retrieval-error-analysis}
\end{table}

We conduct error analysis of predictions from c-REALM + MP classifier in the detection subtask.
The authors annotate 50 false positive instances and 50 false negative instances on the validation set, sampled uniformly at random, and categorize the cause of errors.

Results are reported in in Table \ref{tab:detection-retrieval-error-analysis}.
Overall, most errors are due to failure in retrieval---although the model often successfully retrieves passages that are on right topic, it fails to retrieve passages that contain enough information about whether the backgrounded presuppositions are true or false. This issue is more prominent in false negatives, likely because finding that the presupposition is true requires significantly more exhaustive search than noticing that the presupposition is false (since in most cases, whether the presupposition is correct is not explicitly mentioned).

Secondly, there are 8\% cases of labeling error in false positive cases. Note that for false positive cases, we did not distinguish between direct evidence and indirect evidence because there is no clear definition of ``evidence'' for non~\fp~cases.
For false negative cases, if there is a retrieved passage that can directly \textit{answer} the question but the model fails in labeling, we consider this as a classification error rather than retrieving error, because the goal for retrieval is to retrieve passages that could give the answer to the question. The model also suffers more to classify given indirect evidence that requires reasoning, which matches our intuition. 

Inherent disagreement and inconsistency issues contribute to the rest of the errors. We found that minor part (4\%, 2\% for false positive and false negative cases, respectively) of the error due to the inherent disagreements in labeling ambiguity. We also found that 2\% of the false positive errors are due to whether the \fp~is critical to satisfy users information need based on the question. Furthermore, we also found that 14\% of the errors are due to the comment and the retrieved passages are not inconsistent, causing label mismatch, however, they are not a labeling error because our ground truth annotator is not presented with additional passages when annotating. Example of these inconsistencies can be found in Table \ref{tab:retrieval-inconsistencies-examples}. 

\begin{table}[!t]
    \centering \footnotesize
    \begin{tabular}{p{0.47\textwidth}}
        \toprule
        \Q\ Why aren’t helicopters used to rescue hikers and remove bodies from Mt. Everest?\\
        \C\ The air in too thin and they can't fly that high up. They create lift by pushing air downwards. The higher up you go, the less air pressure you have, the less downward force a helicopter can make to fight gravity.\\
        \textit{\textbf{Retrieved Passage}}: In 2016 the increased use of helicopters was noted for increased efficiency and for hauling material over the deadly Khumbu icefall. In particular it was noted that flights saved icefall porters 80 trips but still increased commercial activity at Everest...\\
        \textit{\textbf{Original Label}}: No \fp.\\
        \textit{\textbf{Model Prediction}}: \fp.\\
        \mycomment\ Helicopter is used to rescue people from Mt. Everest according to the passages, but the comment does not point this \fp~out.\\
        \midrule
        \Q\ How does alcohol effect blood sugar in people with Diabetes?\\
        \C\ ...\textbf{Short answer: straight alcohol doesn’t affect me at all (vodka, whiskey, etc).} I can drink without any noticeable effect on my blood sugar levels (and I have a continuous glucose monitor so I can literally see the effect or lack thereof)...\\
        \textit{\textbf{Retrieved Passage}}:The intake of alcohol causes an initial surge in blood sugar, and later tends to cause levels to fall. Also, certain drugs can increase or decrease glucose levels...\\
        \textit{\textbf{Original Label}}: \fp.\\
        \textit{\textbf{Model Prediction}}: no \fp.\\
        \mycomment\ The comment says that straight alcohol does not affect blood sugar change, but the retrieved passage says otherwise. \\
        \bottomrule
    \end{tabular}
    \caption{Inconsistencies between comment and retrieved passages by c-REALM from Wikipedia. 
    }\label{tab:retrieval-inconsistencies-examples}
\end{table}

\subsection{Discussion on inherent ambiguity and inconsistency on the web}
\label{app:analysis-human-errors}

\begin{table*}[!t]
    \centering \footnotesize
    \begin{tabular}{p{0.45\textwidth}}
        \toprule
        \multicolumn{1}{c}{\textbf{Inherent disagreement: Ambiguity}}\\
        \Q\ How are paintings or museum art classified as masterpieces? Some look like paint scratches and yet they are classics. Why is that?\\
        \C\ Great art isn’t just technical skill. Though a lot of those paintings that may look easy usually aren’t. But it’s also about having an original idea and expressing yourself in a way the world hasn’t seen before. (...)\\
        \mycomment\ Whether the question made the implicit presupposition that a painting or an art is considered a masterpiece only depends on the technical skill exist in the question is debatable before looking at the comment. \\\\
        
        \Q\ Why is Hydrogen so common on Earth and Helium quite rare?\\
        \C\ Hydrogen is highly reactive, it bonds to oxygen, forming water. Water is quite dense, even as a vapor, and is therefore quite durable in the atmosphere. Helium is a noble gas and nearly perfectly inert. Being unbound to any heavier elements, it quickly rises to the top of the atmosphere and is lost to space by various mechanisms. Hydrogen is lost over time, but only slowly.\\
        \mycomment\ There is an \fp~if the question is asking about hydrogen gas, as indicated by \href{https://www.nationalgrid.com/stories/energy-explained/what-is-hydrogen}{this webpage} that hydrogen gas is very rare on earth. However, if they are asking about hydrogen atom, then there is no \fp.\\
        \midrule
        \multicolumn{1}{c}{\textbf{Inherent disagreement: Criticalness}}\\
        \Q\ Why do things look darker when wet?\\
        \C\ You know how when you vacuum the floor it makes those different colored lines? If you look closely the darker colored parts of the carpet are laying down and the lighter colored parts are standing up. Many things that are dry have little hairs or rough surfaces that are basically standing up like little mirrors to reflect light. When these get knocked over or filled with water they can't reflect as well. A couple of damp riddles 1. What gets wetter as it dries 2. What gets darker as it dries Hint: An owlet and naipt\\
        \bottomrule
    \end{tabular}\vspace{-.1em}
    \begin{tabular}{p{0.45\textwidth}}
    \toprule
        \multicolumn{1}{c}{\textbf{Information on the web being inconsistent}}\\ 
        \Q\ Why do 78\% of Americans live paycheque to paycheck?\\
        \textit{\textbf{\href{https://www.cnbc.com/2022/06/27/more-than-half-of-americans-live-paycheck-to-paycheck-amid-inflation.html}{News on 06/07/2022}}}: 58\% Americans are living paycheck to paycheck.\\
        \textit{\textbf{\href{https://www.forbes.com/sites/zackfriedman/2019/01/11/live-paycheck-to-paycheck-government-shutdown/?sh=e6204954f10b}{News on 01/11/2019}}}: 78\% Americans are living paycheck to paycheck.\\\\
        \Q\ Why do pandas eat bamboo and when did they stop eating meat?\\
        \C\ Everyone so far has gone with the "pandas are so dumb" response so let me give you a proper answer. For a start, evolution is not a intelligent or forward thinking process. Bamboo may be low in energy, but it is abundant, grows quickly and not many other animals eat it. So for any animal that can evolve to consume it, there's an open niche there to be taken. Now I'll admit pandas aren't the best creature at surviving, but they don't really need to be. They live in an environment with abundant food and no predators or competitors, so all they need to do is sit around eating bamboo, expend minimal energy and produce just enough babies to keep the species going. Now that might not seem very efficient from a human perspective, but actually it's a strategy that works surprisingly well, and pandas were doing absolutely fine until humans came along and started hunting them and destroying their habitat.\\
        \textbf{\href{https://wwf.panda.org/discover/knowledge_hub/endangered_species/giant_panda/panda/what_do_pandas_they_eat/}{WWF:}} But they do branch out, with about 1\% of their diet comprising other plants and even meat. While they are almost entirely vegetarian, pandas will sometimes hunt for pikas and other small rodents.\\
        \textbf{\href{https://www.science.org/content/article/how-pandas-survive-their-bamboo-only-diet}{Science:}} Pandas are one of the world's most fascinating vegetarians. Their digestive systems evolved to process meat, yet they eat nothing but bamboo—all day, every day. A new study reveals how these animals survive on a diet that should kill them.\\
        \bottomrule
    \end{tabular}\vspace{-.1em}
    \caption{Inconsistencies and inherent ambiguity examples for human performance.
    }\label{tab:inconsistencies-examples}
\end{table*}

Table \ref{tab:inconsistencies-examples} display examples for the category ``Inherent disagreement: ambiguity'', ``Inherent disagreement: criticalness'', and ``Information on the web being inconsistent'' from the human evaluation section of Section \ref{subsec:detection-subtask-results} and in Table \ref{tab:human-perf-detection-task}. 

9.1\% of the errors is due to the inherent ambiguity of the criticalness of the \fp, i.e., whether correcting the \fp~is critical to satisfy users' information needed. For example, for the question ``Why do things look darker when wet?'' in  Table \ref{tab:inconsistencies-examples}, although our human rater found \href{https://paintinggal.com/does-paint-dry-darker-or-lighter-what-you-need-to-know/}{evidence on the internet} that there exist things that look darker when dry, which would contradict the presupposition that (all) things look darker when wet, we believe that the question writer is mainly seeking an answer for the reason of the phenomenon that something looks darker when they are wet, and therefore, such \fp is not critical to answer the users original question, and therefore the comment writer does not point it out. 

Note that this is different than the ``exception'' examples mentioned in Table \ref{tab:data-analysis}, as the comment explicitly pointed out the falsehood of the presupposition, and therefore we consider the \fp~as critical to answer the user's information seeking need. 

Furthermore, 22.7\% of the errors in human performance is due to information on the web being inconsistent. For the question ``why do 78\% of Americans live paycheque to paycheck?'',  \href{https://www.cnbc.com/2022/06/27/more-than-half-of-americans-live-paycheck-to-paycheck-amid-inflation.html}{News on 06/07/2022} points out that 58\% Americans lives paycheck to paycheck, while \href{https://www.forbes.com/sites/zackfriedman/2019/01/11/live-paycheck-to-paycheck-government-shutdown/?sh=e6204954f10b}{News on 01/11/2019} pointed out that 78\% of Americans live paycheck to paycheck. For the question ``why do pandas eat bamboo and when did they stop eating meat?'', creditable sources such as World Wide Fund(WWF) and Science says differently about whether panda eat meat or not. 

34.1\% of the errors are due to ambiguity, as analyzed in Section~\ref{subsec:data-analysis}.

\begin{table*}[!t]
    \centering \footnotesize
    \begin{tabular}{p{0.99\textwidth}}
    \toprule
        \multicolumn{1}{c}{\colorbox{gray!30}{\textbf{\texttt{Inputs given to the human raters}}}
        } \\
        \textbf{\em{Question:}} Why do prosecuters/courts seek/sentence prison time greater than the expected lifespan of the offender (i.e. 150 years in prison)? Why not simply sentence those criminals to 'life' in prison instead?\\
        \textbf{\em{Comment:}} Sentencing options are written into state laws. Life in prison is different in state laws than 150 years. Some of it comes into play with the "cruel and unusual punishment" clause in the Constitution too. Life in prison may not be ``cruel and unusual'' for a murder sentence, but it might be for, say, child sex trafficking. But if you trafficked 10 kids and the sentence is 15 years for each one, you get an effective life sentence that will also stand up, Constitutionally, against a ``cruel and unusual punishment'' defense.\\
        \midrule
        \multicolumn{1}{c}{\colorbox{gray!30}{\textbf{\texttt{Outputs human raters rate}}}
        } \vspace{.3em} \\
        \multicolumn{1}{c}{\textbf{Reference}} \\
        \textbf{\em{Presupposition:}} It does not make sense to sentence a person to 150 years in prison if they can't live that long anyways, prosecutors should use the life in prison sentence instead.\\
        \textbf{\em{Correction:}} The defendant can argue the life in prison sentence as cruel and unusual, so the actual year sentence is better to give than the alternative. \vspace{-.3em} \\
        \multicolumn{1}{c}{\textbf{\NLITrack, Dedicated}} \\
        \textbf{\em{Presupposition:}} Penalties should be able to be sentenced to life in prison.\\
        \textbf{\em{Correction:}} Life in prison is different in state laws than 150 years in prison. \\
        \multicolumn{1}{c}{\textbf{\NLITrack, Unified}} \\
        \textbf{\em{Presupposition:}} If a criminal is sentenced to life in prison, they should be sentenced to life in prison.\\
        \textbf{\em{Correction:}} It is not the case that if a criminal is sentenced to life in prison, they should be sentenced to life in prison. \\
        \multicolumn{1}{c}{\textbf{Main, Dedicated}} \\
        \textbf{\em{Presupposition:}} Penalties should be able to be imposed on criminals for life.\\
        \textbf{\em{Correction:}} The longer the sentence, the more likely the prosecution will seek to sentence the offender to life in prison. \\
        \multicolumn{1}{c}{\textbf{Main, Unified}} \\
        \textbf{\em{Presupposition:}} Prosecutor's should seek prison time greater than the expected lifespan of the offender.\\
        \textbf{\em{Correction:}} It is not the case that prosecutor's should seek prison time greater than the expected lifespan of the offender.\\
        \bottomrule
    \end{tabular}
    \caption{An example of the input and the output human raters are given for the human evaluation of the writing subtask. Note that human raters are not given which output is a reference or from which system.
    }\label{tab:writing-examples}
\end{table*}

\subsection{Details of the writing subtask}
\label{app:writing-subtask-details}

Example generations from each system are shown in
Table~\ref{tab:writing-examples}.
As we discuss in Section~\ref{subsec:writing-subtask-results}, the unified model strongly prefers to generate the correction with prefix ``It is not the case that'' even though we do not restrict it to do so at inference time.

\paragraph{Details in Human Evaluation.}
We conduct human evaluation for 200 questions on the test data.
We recruit two student evaluators, who independently evaluate the presupposition and the correction given a question and its comment.
They evaluated five outputs, including the reference in the data as well as generations from four systems in Section~\ref{subsec:writing-subtask-baselines}: the dedicated and the unified model from the \NLITrack\ and the main track, respectively.

We design detailed evaluation scheme, hold a 1 hour in-person tutorial session for human evaluators to be familiarized with the evaluation task. %
In particular, each output is rated based on four aspects as follows.

\vspace{.5em}
\textbf{Fluency} measures the fluency of the generated text, mainly whether it have repetitions, spelling errors or grammatical errors, or gibberish. %
\vspace{-.3em}
\begin{itemize}%
    \setlength\itemsep{-0.1em}
    \item[\textbf{0:}] Generated text have fluency errors. 
    \item[\textbf{3:}] Generated text is free of fluency errors.
\end{itemize}

\vspace{.1em}
\textbf{Presupposition} evaluates whether the generated presupposition is the valid one in the question and whether it is factually false according to the comment.
\vspace{-.3em}
\begin{itemize}%
    \setlength\itemsep{-0.1em}
    \item[\textbf{0:}] The presupposition is invalid, i.e., does not exist in the question.
    \item[\textbf{1:}] The presupposition is valid, e.g., exists in the question, but it is not factually false.
    \item[\textbf{3:}] The presupposition is valid and is factually false.
\end{itemize}

\vspace{.1em}
\textbf{Correction} evaluates whether the generated correction provides the valid correction to the presupposition based on the comment with no hallucinated information, and provide enough justification (rather than simply being a negated presupposition). The former considers correctness (precision of the information), while the latter considers adequacy (recall of the information)
\vspace{-.3em}
\begin{itemize}%
    \setlength\itemsep{-0.1em}
    \item[\textbf{0:}] The correction is wrong based on the comment, or the correction is hallucinated.
    \item[\textbf{1:}] The correction is correct based on the comment, but no additional information is provided to justify, or is a simple negation of the presupposition.
    \item[\textbf{2:}] The correction is correct based on the comment, but misses some details to fully justify the falsehood of presupposition.
    \item[\textbf{3:}] The correction is correct and provide enough information to justify the falsehood of presupposition.
\end{itemize}

\vspace{.1em}
\textbf{Consistency} requires the generated assumption and correction should be on the same topic and negate each other. %
\vspace{-.3em}
\begin{itemize}%
    \setlength\itemsep{-0.1em}
    \item[\textbf{0:}] The presupposition and correction are not about the same topic.
    \item[\textbf{1:}] The presupposition and correction are on the same topic, but they are not negating each other, or the negation is not explicit.
    \item[\textbf{3:}] The presupposition and correction are consistent: are on the same topic and negate each other.
\end{itemize}

The evaluators are paid with 17 USD/hour. 
See Section~\ref{subsec:writing-subtask-human} for the results and discussion.